\documentclass[letterpaper, 10 pt, journal, twoside]{format/IEEEtran}

\IEEEoverridecommandlockouts    
\usepackage{glossaries}

\newacronym{mav}{MAV}{Micro Aerial Vehicles}
\newacronym{uav}{UAV}{Unmanned Aerial Vehicle}
\newacronym{ovc}{OVC}{Open Vision Computer}
\newacronym{lidar}{LiDAR}{Light Detection and Ranging}
\newacronym{vio}{VIO}{Visual-Inertial Odometry}
\newacronym{gpgpu}{GPGPU}{General-Purpose Graphics Processing Unit}
\newacronym{ugv}{UGV}{Unmanned Ground Vehicle}
\newacronym{uwb}{UWB}{Ultra Wideband}
\newacronym{svm}{SVM}{Support Vector Machine}
\newacronym{fcn}{FCN}{Fully Convolutional Network}
\newacronym{cnn}{CNN}{Convolutional Neural Network}
\newacronym{loam}{LOAM}{LiDAR Odometry and Mapping}
\newacronym{sloam}{SLOAM}{Semantic LiDAR Odometry and Mapping}
\newacronym{slam}{SLAM}{Simultaneous Localization and Mapping}
\newacronym{iot4ag}{IoT4Ag}{NSF Engineering Research Center for the Internet of Things for Precision Agriculture}
\newacronym{grasp-lab}{GRASP Lab}{the General Robotics, Automation, Sensing and Perception Laboratory}
\newacronym{jps}{JPS}{Jump Point Search}
\newacronym{ukf}{UKF}{Unscented Kalman Filter}
\newacronym{sam}{SAM}{Smoothing and Mapping}
\newacronym{slc}{SLC}{Semantic Loop Closure}
\newacronym{icp}{ICP}{Iterative Closest Point}
\newacronym{imu}{IMU}{Inertial Measurement Unit}
\newacronym{tsdf}{TSDF}{Truncated Signed Distance Field}
\newacronym{esdf}{ESDF}{Euclidean Signed Distance Field}
\newacronym{sdf}{SDF}{Signed Distance Field}
\newacronym{rrt}{RRT}{Rapidly Exploring Random Tree}
\newacronym{fpv}{FPV}{First-person View}
\newacronym{dnn}{DNN}{Deep Neural Network}
\newacronym{igpred}{IGPred}{Information Gain Prediction}
\newacronym{csqmi}{CSQMI}{Cauchy-Schwarz Quadratic Mutual Information}
\newacronym{nbv}{NBV}{Next Best View}
\newacronym{vae}{VAE}{Variational Autoencoder}
\newacronym{tsp}{TSP}{Travelling Salesman Problem}
\newacronym{bgsm}{BGSM}{Behavior Guidance State Machine}
\newacronym{pca}{PCA}{Principal Component Analysis}
\newacronym{aspp}{ASPP}{Atrous Spatial Pyramid Pooling}
\newacronym{swap}{SWaP}{Size Weight and Power}
\newacronym{soi}{SoI}{Semantic Object of Interest}
\newacronym{aoi}{AoI}{Area of Interest}
\newacronym{cop}{COP}{Correlated Orienteering Problem}
\newacronym{ig}{IG}{Information Gain}

\usepackage{tabularx}
\usepackage{multirow, multicol, makecell}
\usepackage{array}
\newcolumntype{P}[1]{>{\centering\arraybackslash}p{#1}}
\newcolumntype{M}[1]{>{\centering\arraybackslash}m{#1}}
\usepackage{booktabs}
\newcolumntype{N}{>{\centering\arraybackslash}m{.5in}}
\newcolumntype{G}{>{\centering\arraybackslash}m{2in}}

\newcommand\revise[1]{\textcolor{black}{#1}}
\newcommand\revisefinal[1]{\textcolor{black}{#1}}

\let\labelindent\relax

\usepackage[table,usenames,dvipsnames]{xcolor}      
\usepackage{extarrows}                              
\usepackage{enumitem}
\usepackage{wrapfig}

\usepackage{amsmath,amssymb,amsfonts,amsthm,dsfont, bm} 
\usepackage{mathtools}
\usepackage{algorithm,algorithmicx,listings}        
\usepackage[noend]{algpseudocode}			              
\makeatletter
\def\BState{\State\hskip-\ALG@thistlm}
\makeatother
\DeclarePairedDelimiter\abs{\lvert}{\rvert}%
\DeclarePairedDelimiter\norm{\lVert}{\rVert}%
\makeatletter
\let\oldabs\abs
\def\abs{\@ifstar{\oldabs}{\oldabs*}}
\let\oldnorm\norm
\def\norm{\@ifstar{\oldnorm}{\oldnorm*}}
\makeatother

\usepackage{graphicx}
\usepackage{subcaption}
\usepackage{textcomp}
\usepackage[font=footnotesize]{caption}
\usepackage[breaklinks=true, colorlinks, bookmarks=true, citecolor=Black, urlcolor=Violet,linkcolor=Black]{hyperref}

\usepackage[normalem]{ulem}
\usepackage[utf8]{inputenc}
\usepackage[english]{babel}
\usepackage{hyperref}
\hypersetup{
    colorlinks=true,
    linkcolor=blue,
    filecolor=magenta,      
    urlcolor=cyan,
}

\newcommand*{\sumSr}[2][{}]{\smashoperator[r]{\sum_{#2}^{#1}}}

\DeclareMathAlphabet\mathbfcal{OMS}{cmsy}{b}{n}

\usepackage[space, compress, sort]{cite}

\newtheorem*{assumption*}{Assumption}

\newtheorem*{problem*}{Problem}

\usepackage{lipsum}
\usepackage[capitalise]{cleveref}

\usepackage{tikz}
\newcommand\copyrighttext{%
  \footnotesize \textcopyright 2024 IEEE. Personal use of this material is permitted.
  Permission from IEEE must be obtained for all other uses, in any current or future
  media, including reprinting/republishing this material for advertising or promotional
  purposes, creating new collective works, for resale or redistribution to servers or
  lists, or reuse of any copyrighted component of this work in other works.}
\newcommand\copyrightnotice{%
\begin{tikzpicture}[remember picture,overlay]
\node[anchor=south,yshift=7pt] at (current page.south) {\fbox{\parbox{\dimexpr\textwidth-\fboxsep-\fboxrule\relax}{\copyrighttext}}};
\end{tikzpicture}%
}

\crefname{section}{Sec.}{Secs.}
\Crefname{section}{Sec.}{Secs.}

\begin{document}

\title{3D Active Metric-Semantic SLAM}

\markboth{IEEE Robotics and Automation Letters. Preprint Version. Accepted January, 2024} 
{TAO \MakeLowercase{\textit{et al.}}: 3D Active Metric-Semantic SLAM}  

\author{Yuezhan Tao$^{*}$, Xu Liu$^{*}$, Igor Spasojevic, Saurav Agarwal and Vijay Kumar %
\thanks{Manuscript received: September 13, 2023; Revised December 8, 2023; Accepted January 24, 2024.}
\thanks{This paper was recommended for publication by Editor Tetsuya Ogata upon evaluation of the Associate Editor and Reviewers' comments. This work was supported by TILOS under NSF Grant CCR-2112665, IoT4Ag ERC under NSF Grant EEC-1941529, the ARL DCIST CRA W911NF-17-2-0181, and ONR Grant N00014-20-1-2822.}%
\thanks{$^{*}$Equal Contribution. All authors are with GRASP Laboratory, University of Pennsylvania {\tt\footnotesize\{yztao, liuxu, igorspas, sauravag, kumar\}@seas.upenn.edu}.} %
\thanks{Digital Object Identifier (DOI): 10.1109/LRA.2024.3363542.}
}

\maketitle
\copyrightnotice

\begin{abstract}
In this letter, we address the problem of exploration and metric-semantic mapping of multi-floor GPS-denied indoor environments using \gls{swap} constrained aerial robots. 
Most previous work in exploration assumes that robot localization is solved. 
However, neglecting the state uncertainty of the agent can ultimately lead to cascading errors both in the resulting map and in the state of the agent itself. 
Furthermore, actions that reduce localization errors may be at direct odds with the exploration task. 
We develop a framework that balances the efficiency of exploration with actions that reduce the state uncertainty of the agent. 
In particular, our algorithmic approach for active metric-semantic SLAM is built upon sparse information abstracted from raw problem data, to make it suitable for \gls{swap}-constrained robots.
Furthermore, we integrate this framework within a fully autonomous aerial robotic system that achieves autonomous exploration in cluttered, 3D environments.
From extensive real-world experiments, we showed that by including \gls{slc}, we can reduce the robot pose estimation errors by over 90\% in translation and approximately 75\% in yaw, and the uncertainties in pose estimates and semantic maps by over 70\% and 65\%, respectively.
Although discussed in the context of indoor multi-floor exploration, our system can be used for various other applications, such as infrastructure inspection and precision agriculture where reliable GPS data may not be available. Experiment videos, code, and more details can be found on our project page: \url{https://tyuezhan.github.io/3D_Active_MS_SLAM/}


\end{abstract}
    \section{Introduction}
\label{sec:intro}

\IEEEPARstart{M}{any}  real-world applications require the construction of accurate metric-semantic maps of \textit{a priori} unknown 3D environments. Unlike traditional maps that are concerned only with geometric information in the environment, metric-semantic maps encode both geometric and semantic information. Semantic objects provide a sparse but informative representation of the environment. In addition to benefiting robot navigation, they also provide actionable information for humans, e.g. they aid estimation of yield in agriculture or inventory in factories.

Due to the remarkable progress in deep learning during the past decade, extracting semantic information from the environment, such as object detection or scene classification, can be achieved with off-the-shelf pre-trained neural network models. As a result, we have seen many significant advances in metric-semantic SLAM\cite{chen2020sloam, yang2019cubeslam, salas2013slam++, bowman2017probabilistic, nicholson2018quadricslam, rosinol2021kimera}. 

Autonomous exploration has been widely studied and various approaches and systems have been proposed~\cite{frontier-yamauchi1997frontier, 2d2019semanticIndoorExp, KelseyIG, topo2020indoor}. With the increase in computing power and the emergence of UAVs, recent work has been focused on expanding the planning space into 3D domains~\cite{shen2012indoor, NBVexplore2016, mixed2017FrontierTSP, zhou2021fuel}.

\begin{figure}[t!]
        \centering
        \includegraphics[trim=50 75 0 0, clip, width=0.9\columnwidth]{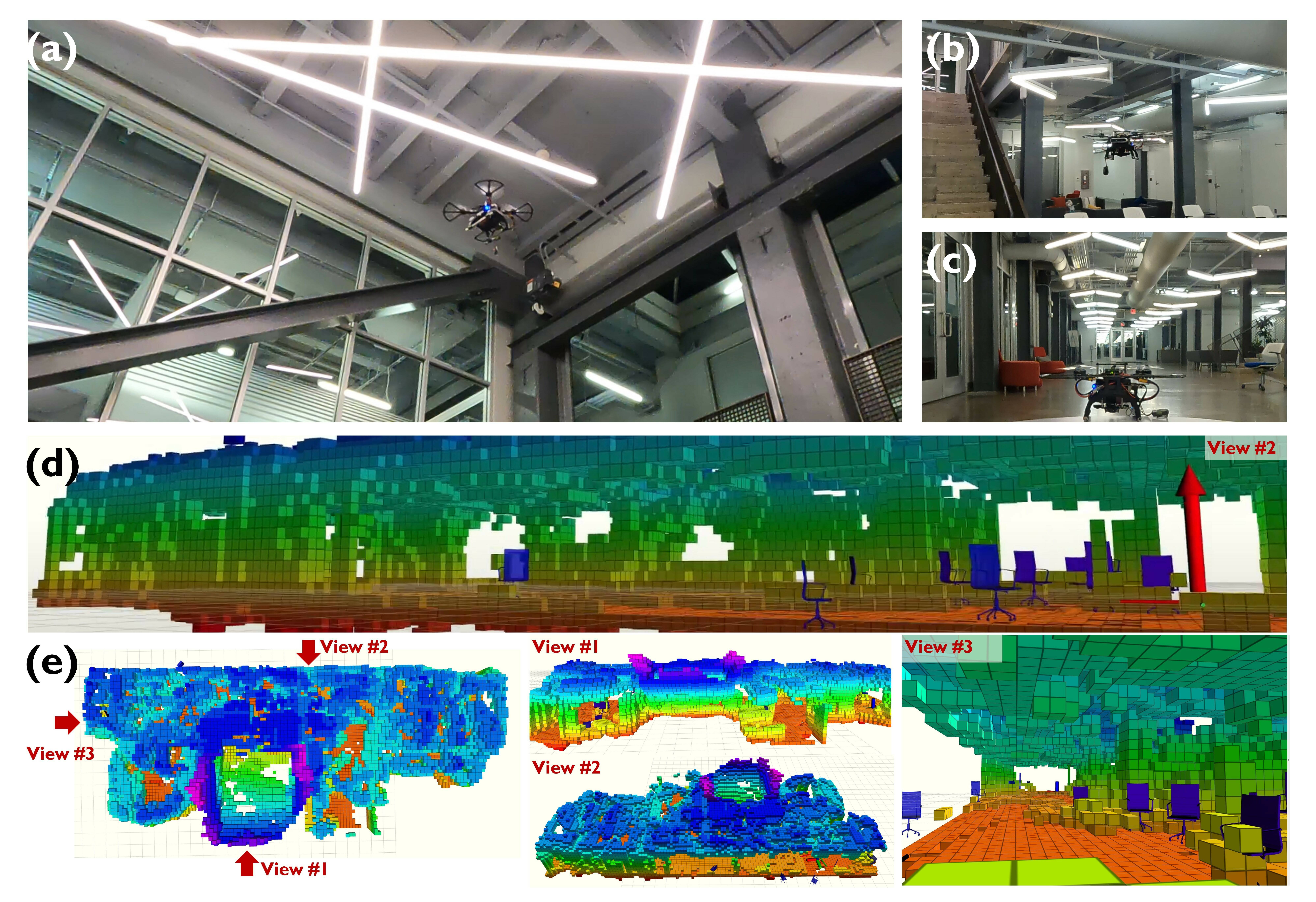}
                \vspace{-0.01in} 
        \caption{\textbf{Falcon 250 UAV exploring a multi-floor environment}. The robot explores the first (b-c) and second (a) floors, while constructing a metric-semantic map (d-e) in real time. Our framework enables efficient 3D exploration and accurate metric-semantic mapping.}
        \label{fig:title-figure-platform}
                \vspace{-0.1in}     
\end{figure}

However, very few of prior works considered the problem of exploration in metric-semantic maps, or active metric-semantic mapping.
Even those that do consider active metric-semantic mapping \cite{liuicra23, asgharivaskasi-atanasov-2022active-semantic-mapping-Octree, placed-atanasov-carlone-2022survey-active-slam}, they decouple the active mapping problem and the localization problem. This is suboptimal, especially when robots have noisy vision-based sensing. While the robot navigates in the environment, the \gls{vio} system inevitably accumulates drift. 
Such errors will eventually lead the robot to deviate from the desired path, resulting in erroneous mapping results and unsafe behaviors.

Motivated by this gap, in this paper, we present a unified framework that addresses the challenge of concurrent exploration, localization, and metric-semantic mapping. \revise{The contributions of the paper consists of:
\begin{enumerate}
    \item An active Semantic Loop Closure (SLC) module and an \gls{slc} algorithm. The active \gls{slc} module generates and evaluates \gls{slc} candidates with a sparse but semantically meaningful representation of the environment. The \gls{slc} algorithm builds upon this representation to detect loop closures and estimate relative pose transformations. 
    \item A framework that trades off exploration and exploitation. The former is \revisefinal{modeled as the} \gls{cop}, and the latter is achieved using \gls{slc}-enabled active uncertainty reduction planning. 
    \item A 3D exploration and navigation stack for a fully autonomous UAV with real-time metric-semantic localization and mapping. Extensive real-world experiments in multi-floor indoor environments demonstrate the performance of the proposed system and its core modules. 
\end{enumerate}
To our knowledge, we are the first to develop and demonstrate a framework that enables \gls{swap}-constrained UAVs to actively balance 3D exploration and uncertainty reduction using metric-semantic maps, while operating in multi-floor environments without using any infrastructure. A demo video can be found at: \url{https://www.youtube.com/watch?v=Kb3s3IJ-wNg}}.



    \section{Related Work}
\label{sec:related work}

\subsection{3D Autonomous Exploration}


\revise{%
Several methods have been proposed for path planning for autonomous exploration in 2D domain~\cite{frontier-yamauchi1997frontier, 2d2019semanticIndoorExp, KelseyIG, topo2020indoor}. With the emergence of UAVs, especially multirotor micro UAVs, recent works focus on developing algorithms and systems that could effectively plan and explore the full 3D space. In \cite{shen2012indoor}, 3D frontiers are detected through a stochastic equation-based method. In \cite{NBVexplore2016}, \gls{nbv} is sampled in 3D space to maximize \gls{ig}.}

\revise{%
Most works select exploration waypoint greedily or within a finite horizon, while the \gls{tsp} has been applied to generate non-myopic plans.
In \cite{mixed2017FrontierTSP, zhou2021fuel}, 3D viewpoints around frontiers are sampled, and a global TSP tour is planned throughout the exploration process. 
Since TSP requires the robot to visit \textit{all} viewpoints in the graph, it does not consider the information provided at each viewpoint.
In this paper, we use the \gls{cop}~\cite{YuSR16}, which has three main attributes: (1)~the vertices have rewards associated with them, (2)~there is a correlation of rewards between vertices, and (3)~a budget constraint limits the number of vertices that can be visited.
The COP, which is a generalization of the Orienteering Problem (OP), maximizes the total reward while exploiting the correlation between vertices.
Correlations capture the fact that visiting a vertex may provide information about other nearby vertices.
The TSP, on the other hand, has no notion of rewards, correlations, or budget constraints.
Thus, from a theoretical perspective, along with the existing qualitative and quantitative results~\cite{YuSR16, AgarwalA23}, the COP models the environment more accurately than the OP and TSP.
Hence, the COP is our choice for path planning for 3D autonomous exploration.
}

\subsection{Active Semantic SLAM}

Prior work has investigated the problem of semantic SLAM or metric-semantic SLAM. Metric-semantic SLAM differs from traditional SLAM in that it not only utilizes traditional geometric features, such as points, lines or planes, but also leverages semantic features, such as object classes. 

The benefits of utilizing semantic features in a SLAM framework are twofold: First, it helps robot localization because object-level features are more informative, storage efficient, and robust to viewpoint changes \cite{salas2013slam++, bowman2017probabilistic, yang2019cubeslam}. This is especially beneficial when integrated in real time with autonomous navigation in GPS-denied, unstructured environments \cite{liu2022large}. Second, it offers robots a high-level understanding of the environment. Such advanced perception capabilities allow the robot to perform tasks with semantically meaningful mission specifications, such as actively gathering information on objects of interest \cite{chen2020sloam, asgharivaskasi-atanasov-2022active-semantic-mapping-Octree} or collaboratively surveying the environment to discover objects \cite{miller2022stronger}. 

In light of these benefits, we utilize sparse semantic landmarks in the environment to reduce the uncertainties in robot state estimation during exploration. We achieve this by actively establishing Semantic Loop Closure (SLC). 
Specifically, the objective is to revisit a viewpoint for SLC, at which a cluster of semantic objects has been discovered to reduce the uncertainties in SLAM.
Existing approaches can be found in utilizing semantic maps for passive loop closures \cite{hughes2022hydra, semanticloop23Yu}, or using \revise{geometric observations for opportunistic} loop closure \cite{zhang2022exploration-active-closure-shaojie}. 
We propose to use the semantic maps for active \gls{slc}, which allows the robot to keep track of the environment at a much larger scale, efficiently detect and estimate relative transformations upon loop closures, and optimize pose estimation and semantic mapping accuracy simultaneously.

\section{Problem Specification}
\label{sec:problem formulation}

\begin{figure}[t!]
        \centering
            \includegraphics[trim=0 0 0 10, clip, width=0.48\textwidth]{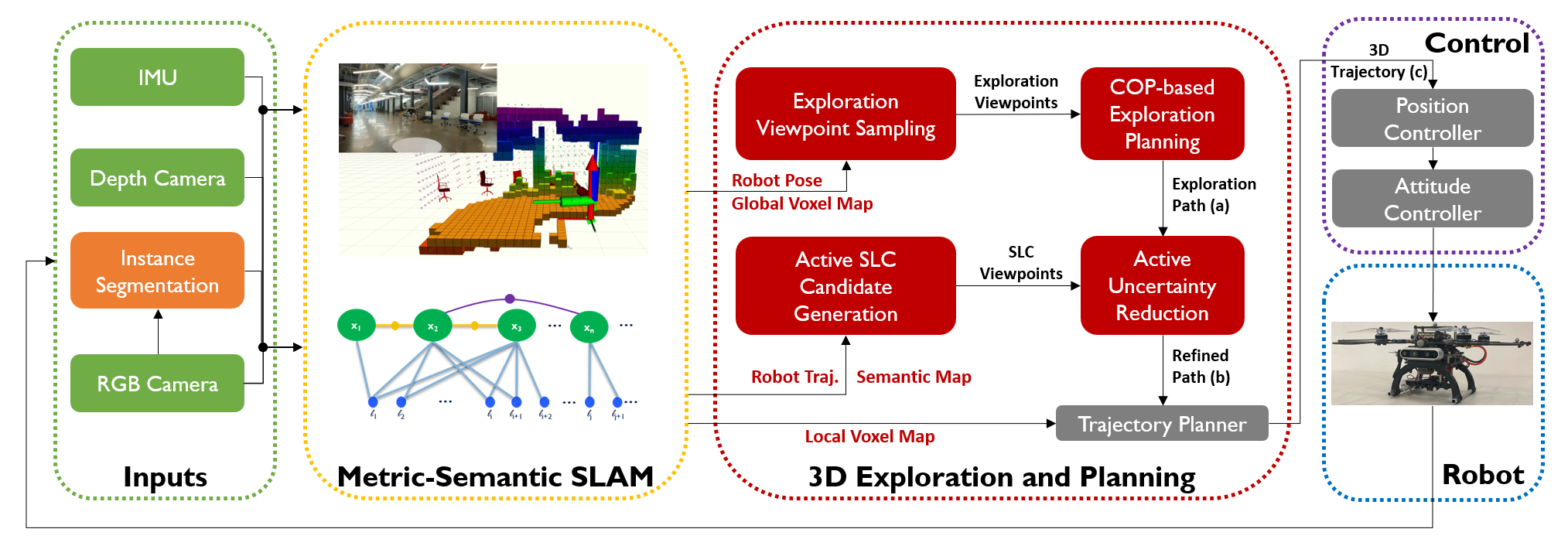}
\vspace{-0.1in}
        \caption{\textbf{System Diagram.} 
        \revise{Our system takes in data from an RGB-D camera and the pose estimates from the VOXL VIO module. Instance segmentation is performed on RGB images with a pre-trained deep neural network (YOLO-V8 \cite{Jocher_YOLO_by_Ultralytics_2023}) model.
The \textit{metric-semantic SLAM} module then takes in these inputs and estimates (1) a global voxel map for sampling exploration viewpoints, (2) a local voxel map for trajectory planning, (3)  optimized robot pose estimates, and (4) a semantic map comprising object landmarks to generate \gls{slc} candidates. Next, a COP-based exploration planning algorithm takes in the exploration viewpoints and plans a long-horizon exploration path (a) consisting of a sequence of viewpoints, which seeks to maximize the Information Gain (IG) given the travel budget. This exploration path is then refined by inserting \gls{slc} viewpoints so that the robot can trade off exploration with uncertainty reduction. The refined path (b) is used to generate goals for the low-level trajectory planning algorithm, which constantly replans dynamically feasible 3D trajectories (c) in the local voxel map. 
}
}
        \label{fig:System Diagram}
 
        \vspace{-0.2in}       
\end{figure}

\revise{
Our goal is to maximize the accuracy of the metric-semantic map of a given region in 3D space within a given exploration budget. 
This requires the robot to 
(1)~efficiently explore the environment, and 
(2)~actively reduce uncertainties in its state estimates and the metric-semantic map.
}

\revise{
In the following sections, we detail the proposed system that enables (1) and (2). From a high level, 
for (1), we model the problem as \gls{cop}, the solution of which provides a long-horizon exploration path. 
For (2), we refine the path by actively establishing \gls{slc}, trading off exploration and uncertainty reduction in metric-semantic SLAM. 
}

\section{System Overview}
\label{sec:system_overview}
We utilize the Falcon 250 platform in this work. This platform, as shown in \cref{fig:title-figure-platform}, carries an Intel Realsense D435i camera, where the RGB images are used for instance segmentation, while the depth images are used for mapping and obstacle avoidance. A Pixhawk 4 Mini flight controller is responsible for low-level attitude control. 

On top of the platform first introduced in \cite{yuezhantao2023seer}, we added a VOXL VIO module \cite{voxlDatasheet}, which outputs six degrees-of-freedom poses at 30 Hz. This, together with the IMU data, is fed into an \gls{ukf} to obtain 150 Hz pose estimates. The platform carries an Intel NUC onboard computer with an i7-10710U processor. The full software stack, including instance segmentation, metric-semantic SLAM, exploration, planning, and control, runs in real time on board. A system diagram and its brief overview are provided and explained in \cref{fig:System Diagram}. In the following sections, we provide detailed explanations of each module of the system.

\section{Metric-Semantic SLAM}
\label{sec: metric-semantic slam}

\subsection{Hierarchical Volumetric Mapping}
\label{sub sec: hierarchical volumetric mapping}

The hierarchical volumetric mapping module maintains two maps with different resolutions: a low-resolution \revise{($f_{gr}$)} global map and a high-resolution \revise{($f_{lr}$)} ego-centric local map. The former is used for frontier \revisefinal{detection, viewpoint sampling, and} COP-based exploration planning, with a size of no less than the experiment region. The latter is used to plan safe local trajectories and has a smaller size \revise{($f_{lx} \times f_{ly} \times f_{lz}$)}.

When the robot receives pose information together with the depth images, ray-casting is conducted to project the readings from the depth images into 3D space, followed by a log-odds-based update on the probabilities of occupancy for all voxels traversed. Map updates are conducted asynchronously for the global map and the local map. The global map is updated with the optimized pose from the metric-semantic SLAM module, while the local map is updated with the estimated pose from the VIO algorithm. 
After each map update, a bounding box enclosing the updated region in the global map is recorded and used by the subsequent frontier detection (see \cref{subsec:frontier_detection_viewpoints}).

\subsection{Semantic SLAM}
\label{subsec: semantic slam, semantic map clustering}
We use a factor graph-based semantic SLAM algorithm. \cref{fig:Active SLAM} shows a close-up view of our factor graph diagram.
Our semantic SLAM algorithm supports different types of objects. It encodes robot pose to object model constraints via customized factors in the GTSAM backend \cite{gtsam,kaess2012isam2}. We refer the readers to our previous work \cite{liuicra23} for details. In this work, we utilize a centroid-based model for the semantic objects in our environment.

The semantic SLAM module takes into account the estimated relative transformation of VIO between two consecutive key poses, i.e., $\mathbf{x}^{vio}_{t} \ominus \mathbf{x}^{vio}_{t-1}$ as the odometry factor, and the estimated centroid locations of the detected objects as the range and bearing factor. 
Due to the sparsity of the semantic map, our factor graph keeps track of historical measurements over the entire mission of the robot, and optimizes the robot poses and object landmarks in a globally consistent manner. 

\subsection{Semantic Loop Closure}
\label{subsec: semantic loop closure}

\revisefinal{The input of the \gls{slc} module consists of the map of} the portion of the environment the robot has explored thus far, and the ``local map'' within its current sensing range.
The module determines whether the robot is currently located at a position it has been in the past, and if so, where. 
It is critical to \revisefinal{minimize the drift in odometry and errors in map estimates.}
The main challenge of detecting loop closures \revisefinal{involves aligning two maps with only partial overlap.}
\revisefinal{If the two maps are encoded in the form of dense point clouds, the problem can be computationally challenging.} 
\revisefinal{Popular existing approaches either} iterate between estimating data association and performing point cloud alignment, or solving convex relaxations of the problem.
Nevertheless, they are either susceptible to being stuck in \revisefinal{suboptimal} local minima, or involve substantial computational resources. In addition, they may fail to find the match due to viewpoint changes.  
\revisefinal{We overcome these difficulties using} semantic information to reduce the size of aligned maps, only focusing on aligning \textit{objects} detected in the two maps. 
This approach (1) \revisefinal{significantly} reduces the size of the alignment problem and (2) is robust to viewpoint changes, allowing us to use an exact exhaustive search algorithm.

To illustrate \gls{slc} in more detail, we are given two sets of points, $\mathcal{A}, \mathcal{B} \subset \mathbb{R}^3$  of cardinalities $|\mathcal{A}| = n$ and $|\mathcal{B}| = m$, respectively.
Points in $\mathcal{A}$ represent centroids of semantic objects in our global map, whereas those in $\mathcal{B}$ represent centroids of semantic objects currently within field of view.
Furthermore, there are subsets $\mathcal{A}_1 \subseteq \mathcal{A}$ and $\mathcal{B}_1 \subseteq \mathcal{B}$ with $|\mathcal{A}_1| = |\mathcal{B}_1| = k$, together with a translation parameter $\mathbf{t} \in \mathbb{R}^3$ and a yaw angle $\psi \in (-\pi, \pi]$, such that
\begin{equation}
    \mathbf{p}_{\sigma_{B}(i)} = \mathbf{R}(\psi)^T( \mathbf{p}_{\sigma_{A}(i)} - \mathbf{t}) ,\quad  1 \leq i \leq k
\end{equation}
for some permutations $\sigma_A, \sigma_B \in Sym(k)$. 
Vector $\mathbf{t}$ encodes the position of the UAV w.r.t. the world frame.
We assume the roll and pitch angles of the robot can be accurately obtained from its IMU.
Therefore, matrix $R(\psi) = \exp([\psi \mathbf{e}_3]_{\times})$, with $\psi$ denoting the yaw angle of the UAV, represents its roll-and-pitch-adjusted orientation w.r.t. the world frame. 
Subsets $\mathcal{A}_1$ and $\mathcal{B}_1$ encode the intersection $\mathcal{A} \cap \mathcal{B}$, while $\sigma_{A}, \sigma_{B}$ encode data association. 
Ultimately, $\mathcal{A}_1, \mathcal{B}_1, k, \mathbf{t}, \psi, \sigma_{A}, \sigma_{B}$ are all \textit{unknown}, and the task of our \gls{slc} module is to compute them. 
Loosely speaking, it is a search-based procedure that works by iterating through the Cartesian product of variations of $\mathcal{A}$ and $\mathcal{B}$ in decreasing order of cardinality, and stores the pair for which the quality of the match is as high as possible. 
We measure the quality of the match between sequences of points $(\mathbf{p}_{\sigma_A(i)})_{i = 1}^{k}$ and $(\mathbf{p}_{\sigma_B(i)})_{i = 1}^{k}$ via the residual function $\mathcal{R}$ defined as 
\begin{equation}
\begin{aligned}
    & \mathcal{R}((\mathbf{p}_{\sigma_A(i)})_{i = 1}^{k}, (\mathbf{p}_{\sigma_B(i)})_{i = 1}^{k}) = \\ & \min_{t \in \mathbb{R}^3, \psi \in (-\pi, \pi]} 
    \frac{1}{k}\sum_{i = 1}^{k} || \mathbf{R}(\psi)^T( \mathbf{p}_{\sigma_A(i)} - \mathbf{t} ) - \mathbf{p}_{\sigma_B(i)} ||_2^2.
\end{aligned}
\end{equation}
Smaller residuals correspond to ``better'' matches.
The residual can be computed analytically by noting that for a fixed yaw angle $\psi$, the optimal translation is given by 
\begin{equation}
    \mathbf{t}^{*}(\psi) = \underbrace{\frac{1}{k} \sum_{i=1}^{k} \mathbf{p}_{\sigma_A(i)}}_{=:\bar{\mathbf{p}}_{\sigma_A}} - \mathbf{R}(\psi) \underbrace{\frac{1}{k} \sum_{i=1}^{k} \mathbf{p}_{\sigma_B(i)}}_{=:\bar{\mathbf{p}}_{\sigma_B}}.
\end{equation}
Defining $\Delta \mathbf{p}_{\sigma_A(i)} = \mathbf{p}_{\sigma_A(i)} - \bar{\mathbf{p}}_{\sigma_A}$, and $\Delta \mathbf{p}_{\sigma_B(i)} = \mathbf{p}_{\sigma_B(i)} - \bar{\mathbf{p}}_{\sigma_B}$, the optimal $\psi$ can be recovered by minimizing the expression $\sum_{i = 1}^{k} || \Delta \mathbf{p}_{\sigma_A(i)} - \mathbf{R}(\psi) \Delta \mathbf{p}_{\sigma_B(i)} ||_2^2$ over $\psi$, which in turn, is equivalent to maximizing $tr(\mathbf{R}(\psi) \sum_{i=1}^{k} \Delta \mathbf{p}_{\sigma_B(i)} (\Delta \mathbf{p}_{\sigma_A(i)})^T)$. 
Defining the matrix 
$M := \sum_{i=1}^{k} \Delta \mathbf{p}_{\sigma_B(i)} (\Delta \mathbf{p}_{\sigma_A(i)})^T) \in \mathbb{R}^{3 \times 3}$, we have that the trace under question equals 
\begin{equation}
    \cos(\psi)(M_{11} + M_{22}) + \sin(\psi)(M_{12} - M_{21})+ M_{33},
\end{equation}
and its maximum value 
\begin{equation}
\sqrt{(M_{11} + M_{22})^2 + (M_{12} - M_{21})^2} + M_{33}
\end{equation}
is attained for 
\begin{equation}
\psi^{*} = \arctan 2(M_{12} - M_{21}, M_{11} + M_{22}).
\end{equation}
The running time of this module is $\mathcal{O}((\min(m,n)+1)! \ 2^{\max(m,n)})$.
Even though the the algorithm is exponential, given that sets $\mathcal{A}$ and $\mathcal{B}$ comprise of a small number of semantic objects instead of dense point clouds, we consider the worst case computational burden to be acceptable in practice. 
\revise{Finally, it is worth noting that by disregarding pairs of variations and combinations of $\mathcal{A}$ and $\mathcal{B}$ that match objects of different classes, the complexity bound above can be reduced further. 
Nevertheless, the speed-up involved depends on the distribution of objects across the different classes, and this is something which is unknown a priori.}

\section{Exploration with Active SLC}

In this section, we introduce the exploration with active \gls{slc} module, which utilizes the metric-semantic maps to generate paths that balance exploration and uncertainty reduction.  
\subsection{Frontier Detection and Exploration Viewpoint Sampling}
\label{subsec:frontier_detection_viewpoints}

We employ the incremental frontier detection and viewpoint sampling module presented in our previous work~\cite{yuezhantao2023seer}. All existing frontiers within the bounding box from the map update will be re-evaluated and removed if observed. New frontiers are detected and clustered. Large clusters are broken down into small ones recursively if they are greater than the desired size, \revise{so that the robot can cover the cluster with the limited sensing range and field of view.} 3D viewpoints are sampled for each frontier cluster following a two-step process. 
In the first step, candidate positions are uniformly sampled around the cluster centroid. For the second step, multiple yaw angles are uniformly sampled at each candidate position. 
Different from~\cite{yuezhantao2023seer}, cell-counting-based \gls{ig} is estimated for all sampled yaw angles without information prediction. The candidate yaw angle with maximum estimated \gls{ig} is selected as the sampled yaw angle and associated with the candidate position. We take the sampled pose with the highest estimated \gls{ig} as the viewpoint for the frontier cluster.

\subsection{COP-based Exploration Planning}

The COP operates on a given complete graph $G=(V, E)$, where $V$ is the set of vertices and $E$ is the set of edges.
A vertex $v \in V$ has a reward $r_v \geq 0$ associated with it, and an edge $(i, j) \in E$ has a travel cost $c_{ij} \geq 0$.
\revise{The edge costs are symmetric, i.e., $c_{ij} = c_{ji}$.}
For exploration planning, the vertices represent the sampled viewpoints, and the edges represent optimal paths between viewpoints.
The reward $r_v$ of a vertex $v$ is the estimated \gls{ig} at the viewpoint $v$, while the edge costs are computed using the $A^*$ path cost between the viewpoints.
Additionally, a correlation function $w(u, v) \in [0, 1]$ is defined for each pair of vertices $u, v \in V$, which measures the correlation between the rewards of the two vertices.
We compute the correlation between two vertices as the percentage of the overlap of the two viewpoints, \revise{assuming they are occlusion-free}.
The correlation function is symmetric in this case, i.e., $w(u, v) = w(v, u)$.

The goal of COP is to find a tour (or path)~$\pi$ that visits a subset of the vertices to maximize the total reward collected while respecting a given budget $B$ on the total travel cost.
Let $x_v\in \{0, 1\}$ denote whether a vertex $v$ has been visited and let $y_{ij}\in \{0, 1\}$ denote whether an edge $(i, j)$ has been traversed from vertex $i$ to vertex $j$ by a tour~$\pi$.
The COP maximizes the total reward:
\begin{equation}
 R(\pi) = \sum_{v\in V} r_v \left(x_v + \omega_v(1-x_v)\right),\\
\end{equation}
subject to the following constraints:
\begin{align}
	\omega_v - \sumSr{u\in V\setminus \{v\}} w(u, v)\, x_u &\leq 0\label{eq:cop:omega}\\
	\sum_{(i, j)\in E} \left(c_{ij}\, y_{ij}  + c_{ji}\,y_{ji}\right)&\leq B\label{eq:cop:budget}\\
			\text{Tour constraints for }&\pi\,\text{\cite{YuSR16,AgarwalA23}}\nonumber\\
   			y_{ij}, y_{ji} \in \{0, 1\},& \quad \forall (i, j)\in E\nonumber\\
			\omega_v \in [0, 1], x_v\in \{0,1\}& \quad \forall v\in V,\nonumber
\end{align}

\begin{figure}[!t]
        \centering
            \includegraphics[trim=10 0 30 15, clip, width=0.45\textwidth]{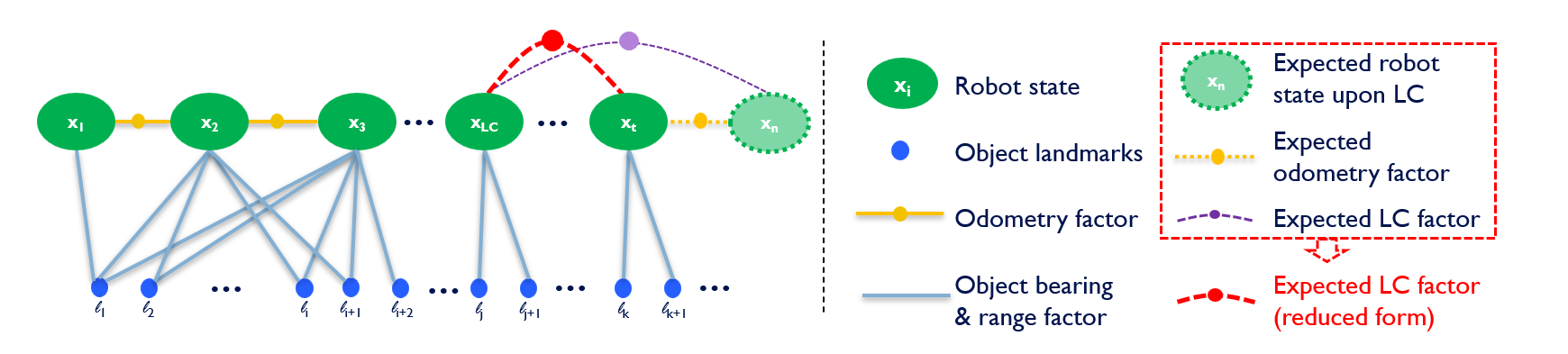}
                \vspace{-0.05in}   
        \caption{\textbf{Active Metric-Semantic SLAM.} The proposed semantic factor graph consists of nodes for both robot poses and object landmarks, and edges that represent odometry constraints, robot-to-object constraints, and semantic loop closure constraints. This graph also illustrates how the virtual factors and nodes are added in the active semantic loop closure step \revise{(see \cref{subsec:active loop closure})}.}

        \label{fig:Active SLAM}
                \vspace{-0.2in}     
\end{figure}

The variable $\omega_v$ models the portion of the reward $r_v$ that is collected by vertices other than the vertex~$v$~\eqref{eq:cop:omega}. 
Similar to~\cite{AgarwalA23}, we permit the sum of correlations, the second term in \eqref{eq:cop:omega}, to be greater than one, unlike the original, more restrictive COP formulation~\cite{YuSR16}.
Note that the variable $\omega_v$ will always be either one or the sum of correlations, as it is in the objective function of a maximization problem.
\revise{Although our edge costs and correlation functions are symmetric, the COP formulation allows them to be asymmetric.}
Constraint~\eqref{eq:cop:budget} is the budget constraint, which limits the total travel cost of the tour. 
We heuristically set the budget by estimating the cost from the robot's position to several nearby frontiers and scale it with a constant factor to limit the resulting tour length and reduce computational complexities. 
Tour constraints ensure that the tour has no disconnected subtours and at least one edge connected to a visited vertex is traversed. 
We refer readers to~\cite{YuSR16,AgarwalA23} for details on the tour constraints.

The COP is NP-hard, and the MIQP formulation~\cite{YuSR16} is not suitable for online computation in the exploration problem.
Hence, we use a simplified version of the greedy constructive heuristic algorithm from~\cite{AgarwalA23}.
The algorithm starts with an empty tour, greedily selects a vertex to be added, and computes an efficient tour with the selected vertices.
These steps are iteratively executed until the budget constraint is violated.
The greedy criterion is based on the value of a vertex computed as:
$\text{value}(v) = r_v + \sum_{u\in V\setminus S} r_{u}\,w(v, u) - \sum_{u\in S} r_{u}w(u, v)$, where $S$ is the subset of vertices already selected in the previous iterations. The complexity of the algorithm is $\mathcal{O}(|V|^3)$~\cite{AgarwalA23}. In practice, by selecting a \revise {proper frontier cluster size ($f_{sz}$)}, we can bound the number of viewpoints in the environment to be less than 10, \revisefinal{making it possible to compute an optimal tour for the selected vertices with the Bellman-Held-Karp Algorithm in real time on board.}
The exploration tour is re-planned if either of the following conditions is met: (1) \revise{a fraction ($f_{r1}$) of frontier} changes in the current environment, (2) the percentage of the refined exploration tour (detailed in \cref{subsec: uncertainty minimizing planning}) that has been executed exceeds \revise{a given threshold ($f_{r2}$)}. The second condition is also known as receding-horizon planning.
The COP-based exploration planning module is asynchronous with the rest of the software stack; the robot continues executing the refined tour until a new refined tour is received.

\subsection{Active Semantic Loop Closure}
\label{subsec:active loop closure}
\begin{figure}[t!]
        \centering
            \includegraphics[trim=30 15 0 10, clip, width=0.42\textwidth]{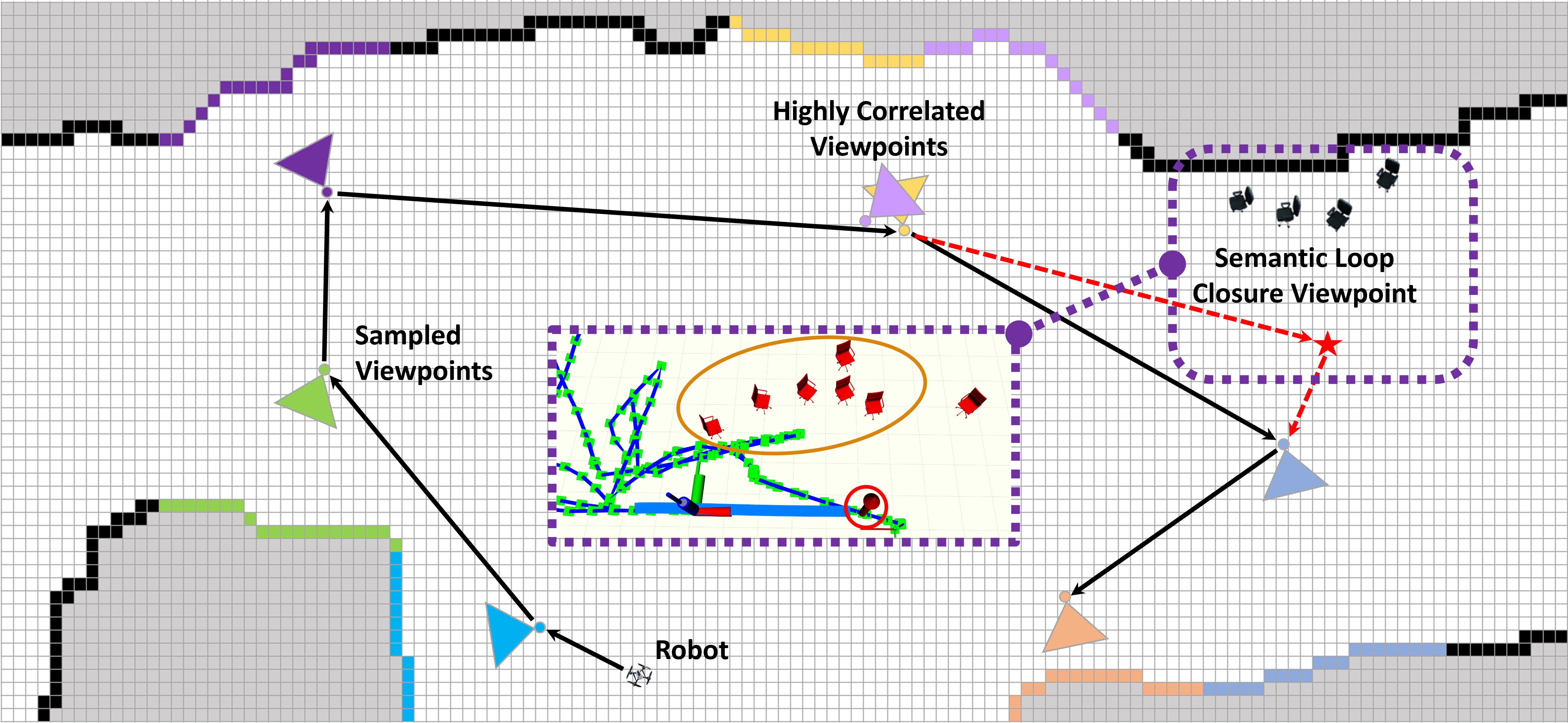}
                    \vspace{-0.05in}  
            \caption{ \textbf{An illustration of exploration with active SLC}. The solid black arrows show the nominal COP-based exploration path. The dashed red arrows highlight the difference between the refined (red) and nominal (black) paths. The active uncertainty reduction planning balances exploration and uncertainty reduction. A pair of active SLC landmark cluster and viewpoint is highlighted in the purple box, in which the orange-circled chairs belong to the cluster, and the red-colored arrow is the \gls{slc} viewpoint (i.e., $\mathbf{x}_{lc}$ in \cref{fig:Active SLAM}).}
                    \vspace{-0.25in}     
    \label{fig:Active SLAM Landmark Cluster}
    \label{fig:ExplorationWithActiveSemanticLoopClosureIllustration}
\end{figure}

While the robot is constructing the metric-semantic map, it needs to generate candidate \gls{slc} submap and viewpoint pairs. 
This is done by first finding the submaps by clustering the semantic landmarks, and then selecting the corresponding viewpoints. 
In the first step, we use the DBSCAN algorithm \cite{ester1996density-dbscan} to cluster the centroids of the semantic landmarks in the Euclidean space, as illustrated in the purple box of \cref{fig:Active SLAM Landmark Cluster}. 
\revise{We obtain valid submaps by choosing clusters with no less than a specific number ($f_{cs}$) of landmarks. Note that} since we have range and bearing measurements from each landmark, with \textit{a priori} unknown data association, we need at least three landmarks to uniquely determine the position and yaw of the robot upon loop closure, as explained in detail in our previous work \cite{spasojevic2023robust}. 
In the second step, for each submap, we need to generate a viewpoint that is reachable, detectable, and informative.
To make it reachable, we choose the viewpoint from the set of key poses (which the robot has reached before) in the factor graph. Next, we limit the choice of the key pose so that any of the landmarks in the submap is within \revise{the sensing range ($f_{sr}$)}. Third, to maximize the possible information gain brought about by the \gls{slc}, the oldest key pose (the pose first added to the factor graph) among all key poses that satisfy the previously mentioned conditions is selected as the viewpoint. 
The robot establishes a loop closure by taking a panorama (by yawing in place) at such an \gls{slc} viewpoint. One example \gls{slc} viewpoint is shown by the red arrow in \cref{fig:Active SLAM Landmark Cluster}. 

Once the loop closure viewpoint-submap pairs are sampled, they will be used in the active uncertainty reduction planning module of our system, as illustrated in \cref{fig:System Diagram}. This module seeks to insert loop closure viewpoints along the COP exploration path, such that the combined \gls{ig} is maximized while respecting the travel budget constraint. Our pseudo-code in \cref{alg:active_closure} further explains this procedure.

An important step is to predict the \gls{ig} brought about by each of the candidate SLC viewpoint-submap pairs. 
We achieve this by adding a virtual factor to the semantic factor graph as illustrated in \cref{fig:Active SLAM}. Conceptually, in this step, we added two factors, an expected odometry factor and an expected loop closure factor. 
The former brings the robot to the loop closure viewpoint ($\mathbf{x}_n$) to establish an \gls{slc} with an existing key pose ($\mathbf{x}_{lc}$) in the graph, and the latter connects $\mathbf{x}_n$ and $\mathbf{x}_{lc}$. Since our loop closure viewpoint is sampled from one of the existing key poses in the graph, $\mathbf{x}_n$ and $\mathbf{x}_{lc}$ are the same nodes. Therefore, the procedure reduces to adding the expected odometry factor between $\mathbf{x}_{t}$ and $\mathbf{x}_{lc}$, with a motion noise scaled by the expected travel distance. For a long-horizon path, we can sequentially perform such operations to evaluate the \gls{ig} for a sequence of actions with multiple SLCs.

This simulates the effect of the robot directly navigating to establish \gls{slc} with its noisy odometry measurements. 
This virtual factor leaves the estimates intact, but alters the covariance matrix of the factor graph. 
We calculate the reduction in the trace of the covariance matrix before and after adding this virtual factor as the \gls{ig} measure: 
    $    IG = tr(\mathbf{\Sigma}_{t}) - tr(\mathbf{\Sigma}_{t+1})$,
where $tr$ denotes the trace of the covariance matrix. Once we evaluate the \gls{ig} along a given path, we remove such virtual factors from the factor graph, so that these virtual factors do not alter the factor graph's estimates.

\subsection{Active Uncertainty Reduction Planning}
\label{subsec: uncertainty minimizing planning}
\subsubsection{Algorithm}

This algorithm converts the planned exploration path from the COP module into a refined path by balancing the \gls{ig} from exploration and uncertainty reduction from \gls{slc}.
Pseudo-code for this module is provided in \cref{alg:active_closure}. 
It iterates through the sequence of viewpoints comprising the COP-based path, at each point evaluating if the robot should actively seek an \gls{slc} before resuming exploration. 
\revise{At every iteration, the non-negative remaining \gls{ig} of the subsequent exploration viewpoint is evaluated by subtracting the correlated information gathered by the viewpoints already present in the refined path from its \gls{ig}.}
\revise{Then, the cost of every \gls{slc} candidate is evaluated via $A^*$ and its \gls{ig} is calculated using the method detailed in \cref{subsec:active loop closure}.
If the budget permits addition of the best \gls{slc} candidate, we further compare the \revise{scaled ($f_{sc}$)} \gls{ig} of the latter with the utility (i.e. cost-benefit index) of the upcoming viewpoint.
The comparison result determines the candidate \gls{slc} candidate should be inserted into the refined exploration path. }
We only allow one \gls{slc} candidate to be inserted between consecutive exploration viewpoints, which also bounds the total running time complexity.

\begin{figure}[t]
\vspace{-.2cm}
\begin{algorithm}[H]
  \scriptsize
  \captionsetup{font=scriptsize}
\caption{Active Uncertainty Reduction Planning}\label{alg:active_closure}
LC: loop closure candidate; BLC: best loop closure candidate; VP: frontier viewpoint;\\
\textbf{Input:} ExpPath \Comment{exploration path from COP}\\
\textbf{Output:} ReExpPath \Comment{\textit{refined} exploration path}
\begin{algorithmic} [1]
\State {ReExpPath $\gets$ [];}
\For{$i \gets 1$ to ExpPath.size()} 
    \State {VP $\gets$ ExpPath(i);}
    \State {VP.IG $\gets$ ComputeRemainingIG(ReExpPath, VP);}
    \For {$j \gets 1$ to LC.size()}
        \State {Estimate Cost \& IG for LC(j);}
        \State {UpdateCurrentBLC;}
    \EndFor
 \revise{   \If {Scale(BLC.IG) $>$ (VP.IG/VP.Cost) and BudgetEnough(BLC)}
        \State\hspace{0.65mm}{Insert BLC to ReExpPath;}
        \State {Insert VP to ReExpPath if BudgetEnough(VP);}
    \ElsIf {BudgetEnough(VP)}
        \State {Insert VP to ReExpPath;}
    \Else 
        \State{return;}
    \EndIf}
\EndFor
\end{algorithmic}
\end{algorithm}
\vspace{-1cm}
\end{figure}

\subsubsection{Compexity Analysis}
\label{sec: complexity analysis}
Suppose that we have $E$ exploration nodes, $C$ loop closure candidates, $S$ semantic landmarks, and $P$ robot poses. 
The outer loop in line 2 requires $\mathcal{O}(E)$ iterations. In each iteration, the dominant cost comes from lines 4 to 7. The computational complexity of line 4 is $\mathcal{O}(E)$. 
The inner loop in line 5 is executed $\mathcal{O}(C)$ times. In every iteration of the inner loop, we run two sub-procedures. The first procedure, which estimates the cost of the \gls{slc} candidate, runs in time required to complete an $A^*$ search - say $T_{A^*}$. The second procedure, which computes the \gls{ig}, runs in time $\mathcal{O}\left(max(P, S)^{1.5}\right)$ \cite{kaess2012isam2}. Collecting the latter, the running \revisefinal{time} of our algorithm is \revise{$\mathcal{O}(E^2 + EC(T_{A^*} + \max(P,S)^{1.5}))$}.

In the case of traditional SLAM, where dense geometric features are used, hundreds of features are tracked for each key pose. 
In this case, $max(P,S) = S$, which is at the order of $100P$ or even larger. 
However, in our case, $S$ will usually be no larger than $P$ since semantic landmarks are sparse.
Thus, $max(P,S)=P$. 
We can further reduce it by, for example, only adding pose nodes whenever we observe a semantic landmark. 
Practically, given $E \ll P$ and $C \ll P$, the complexity of the algorithm is reduced to $\mathcal{O}(T_{A^*} + P^{1.5})$. 
This is essentially doing $A^*$ searches and solving semantic SLAM problem with loop closures multiple times. By searching over a low-resolution map, the $A^*$ search is manageable. Again, since the number of landmarks is much smaller in the semantic SLAM problem than in a traditional SLAM problem, this can be done efficiently online onboard the robot.

\subsection{Drift Compensation and Trajectory Planning}
\label{subsec: low level planning}

Compared to VIO, our semantic SLAM algorithm outputs pose estimates with higher accuracy. However, the smoothness may be sacrificed due to intermittent drift correction induced by \gls{slc}. 
To solve this, we employ a drift compensation module similar to our previous work \cite{liu2022large}. It transforms the next local planning goal from the SLAM reference frame to the odometry reference frame, using the difference between the VIO and semantic SLAM pose estimates.

By this design, the robot's controller, local mapper, VIO, and trajectory planner operate in the odometry reference frame. The exploration planner, semantic SLAM, and global mapper operate in the SLAM reference frame. 
We use~\cite{gcopter} for local trajectory planning and the yaw optimization approach introduced in our previous work ~\cite{yuezhantao2023seer}.

\begin{figure}[t!]
        \centering
        \includegraphics[trim=0 30 0 0, clip, width=0.86\columnwidth]{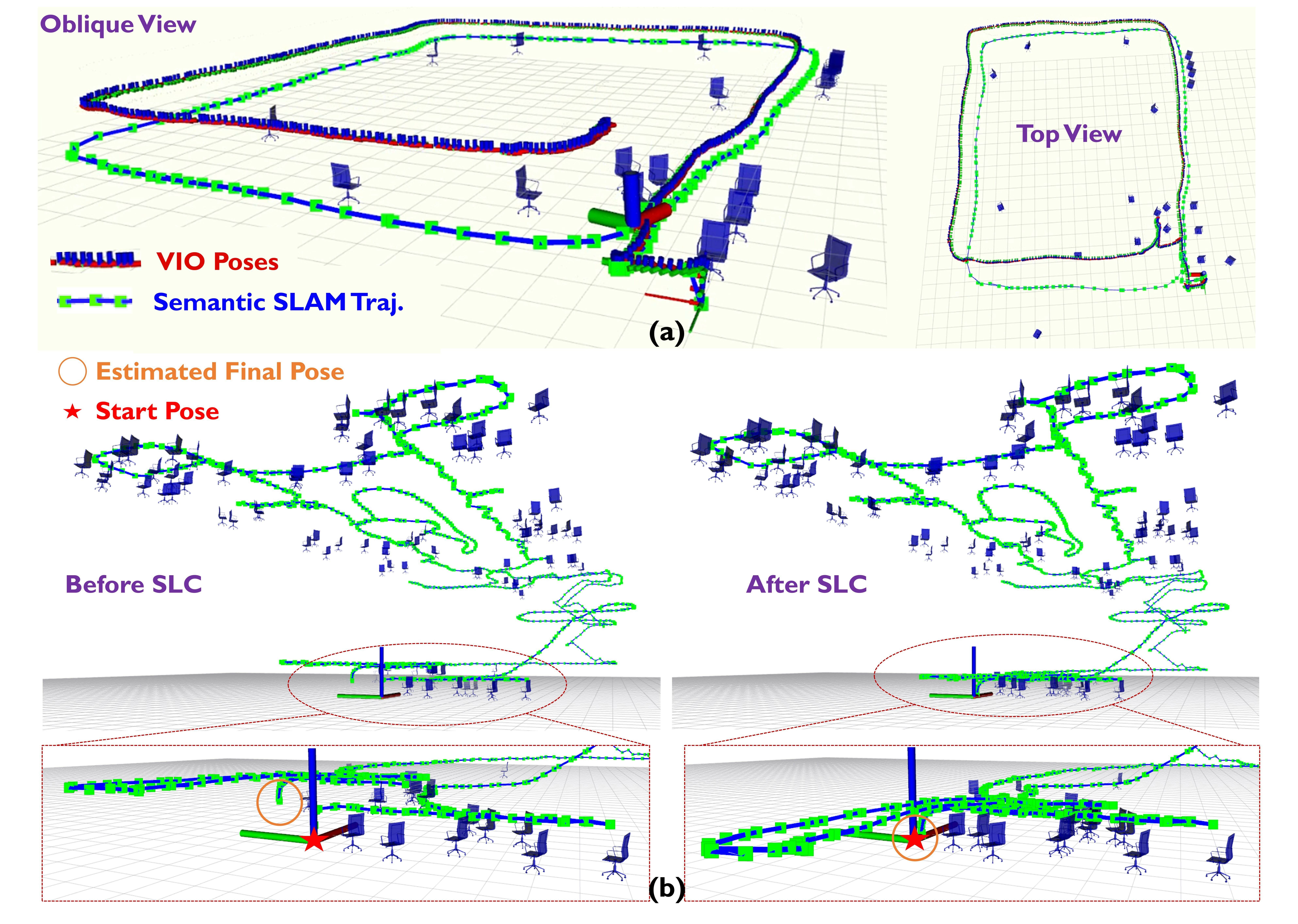}
        \vspace{-0.05in}    
        \caption{\textbf{Robot trajectories and semantic maps with and without \gls{slc} for one-floor (a) and three-floor (b) experiments}. The robot starts and ends at the exact same location. 
        \gls{slc} significantly improves both the semantic map and the robot trajectory. 
        A detailed analysis is provided in \cref{subsec: results-semantic loop closure}.
        }
        \label{fig:qualitative-loop-closure}
                \vspace{-0.25in}     
\end{figure}


\begin{figure*}[t!]
        \centering
            \includegraphics[trim=0 0 10 0, clip, width=0.95\textwidth]{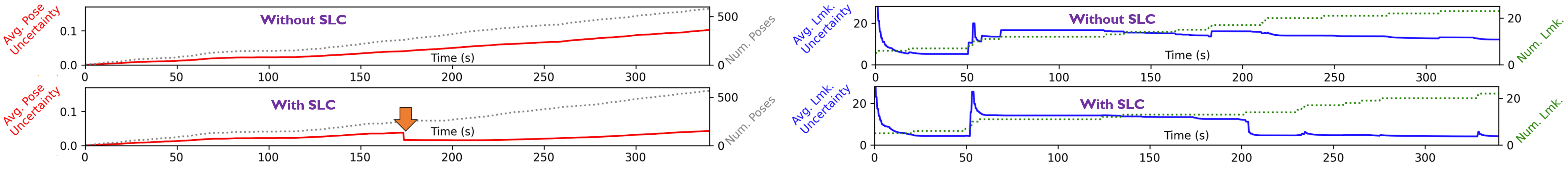}
                    \vspace{-0.1in}
            \caption{\textbf{Uncertainty without (top) and with (bottom) active SLC}. 
            The red and blue lines represent average uncertainties in robot pose and semantic landmarks, respectively. 
            The uncertainty is defined as the trace of covariance matrix in semantic factor graph. The orange arrow shows when \gls{slc} takes place.}
            \label{fig:Uncertainty without (left) and with (right) active loop closure}
                    \vspace{-0.1in}     
\end{figure*}

    \section{Results and Analysis}
\label{sec:results}

To evaluate the efficiency and performance of the entire system and critical modules, we conducted \revisefinal{four} sets of real-world experiments: (1) we evaluated the CPU utilization to empirically estimate the computational requirements of the software stack; (2) we studied the effects of the semantic loop closure module on datasets collected by surveying the building and establishing \glspl{slc} by revisiting the places with clusters of semantic landmarks at the end; (3) we carried out autonomous exploration and metric-semantic mapping experiments where the robot autonomously explored a multi-floor environment; \revisefinal{(4) we benchmarked our system with state-of-the-art SLAM methods.} \revise{
In all experiments, we set $f_{r1}=15\%$,  $f_{r2}=10\%$, $f_{sz}=1.2m$, $f_{sr}=5m$, $f_{sc}=6$, $f_{cs}=4$, $f_{gr}=0.25m$, $f_{lr}=0.1m$, $f_{lx}=15m$, $f_{ly}=15m$, $f_{lz}=4m$.
}
\begin{table*}[!t]
\caption{\textbf{Quantitative results on error reduction in position and yaw estimation.}}
\vspace{-0.15in}
\scriptsize
\begin{center}
\label{tab:benchmark}
\begin{tabular}{
|P{0.12\textwidth}|
P{0.15\textwidth}| 
P{0.15\textwidth} | 
P{0.06\textwidth}  | 
P{0.06\textwidth} | 
P{0.06\textwidth} | 
P{0.06\textwidth} | 
P{0.08\textwidth} | 
}
\hline
{\multirow{2}{*}{\textbf{Mission}}} & \multicolumn{3}{c|}{\textbf{Position Error (m)}} &  \multicolumn{3}{c|}{\textbf{Yaw Error (deg)}} & {\multirow{2}{*}{\textbf{Traj. Len. (m)}}}\\ 
\cline{2-7} 
 & {\textbf{VIO(X/Y/Z)}} & {\textbf{Ours(X/Y/Z)}}  & {\textbf{Reduction}} & {\textbf{VIO}} & {\textbf{Ours}} & {\textbf{Reduction}} &  \\ 
\hline

Loop 1 & 2.50 (-1.95, -1.52, 0.34)  & 0.18 (-0.12, -0.11, 0.08) & \textbf{92.68\%} & 2.96$^{\circ}$ & -1.77$^{\circ}$ & \textbf{39.99\%} & 179.33\\
\hline
Loop 2 & 2.63 (2.36, 0.39, 0.80) & 0.41 (0.02, -0.18, 0.37) & \textbf{83.84\%} & 7.05$^{\circ}$ & -2.69$^{\circ}$ & \textbf{61.82\%} & 454.85\\
\hline
Loop 3 & 4.15 (2.56, 3.16, 0.82) & 0.67 (0.52, 0.42, -0.03) & \textbf{83.78\%} &-12.98$^{\circ}$ & -7.81$^{\circ}$ & \textbf{39.79\%} & 497.83\\ 
\hline

\end{tabular}
\vspace{-0.25in}
\end{center}
\end{table*}

\begin{table}[!t]
\caption{\textbf{Quantitative results on uncertainty reduction (U. Red.) in robot poses and semantic landmarks.}}
\vspace{-0.15in}
\scriptsize
\begin{center}
\label{tab:benchmark_reduction}
\begin{tabular}{
|P{0.04\textwidth}|
P{0.125\textwidth}| 
P{0.06\textwidth} | 
P{0.065\textwidth}  | 
P{0.045\textwidth} | 
}
\hline
{\multirow{2}{*}{\textbf{Mission}}} & \multicolumn{1}{c|}{ \multirow{2}{*}{\parbox{2.5cm}{\centering \textbf{U. Red. of Avg. Pose upon SLC}}}} &  \multicolumn{2}{c|}{\textbf{U. Red. (w/ v.s. w/o SLC)}} & {\multirow{2}{*}{\parbox{1cm}{\centering \textbf{Traj. Len. (m)}}}}\\ 
\cline{3-4} 
 &  &  {\textbf{Avg. Pose}} & {\textbf{Avg. Lmk.}} & \\ 
\hline
Auto 1 & 56.67\%  & 52.06\% & 68.53\% & 227.47 \\
\hline
Auto 2 & 45.72\% & 54.87\% & 26.37\% & 72.62 \\
\hline
Auto 3 & 52.98\%; 14.62\% &70.99\% &  23.93\% & 185.17 \\ 
\hline
\end{tabular}
\vspace{-.75cm}
\end{center}
\end{table}


\subsection{Computational Requirements}
We empirically evaluated the CPU utilization of our system using our UAV's onboard computer as mentioned in \cref{sec:system_overview}. The total CPU utilization is 42.2-53.3\% for the full stack. The majority of the computation is taken by the semantic SLAM front end, which includes an instance segmentation neural network and a point cloud processing module, taking in total $\sim$34\% of the CPU. Note that we used the \revisefinal{medium} version of the YOLO V8, i.e. yolov8m, and we limited the inference rate to 2 Hz. The backend of the semantic SLAM, i.e. the optimization of the factor graph, took 0.88\%. This is an average load, which may include surges when loop closures are triggered. The COP-based exploration module took 0.58\%. The \gls{slc} module utilized 0.23\%.  The rest of the CPU utilization was taken by the remaining modules in our navigation stack, including the voxel mapper, viewpoint sampler, trajectory planner and tracker, state machine, controller, etc. VIO was done on the VOXL board. An important aspect to note is that different modules of our stack execute asynchronously. The delay in one module does not propagate to the other modules.

\subsection{Semantic Loop Closure}
\label{subsec: results-semantic loop closure}

We carried out multiple loop closure experiments inside a cluttered three-story building. 
Two examples are shown in \cref{fig:qualitative-loop-closure}. 
In the one-floor experiment (a), the robot traveled a squared loop on a single floor. 
On the left panel of (a), it is clear that the VIO drifted along the Z direction, which was significantly corrected by the \gls{slc}. 
The right panel of (a) indicates the drift of VIO along X-Y axes, while the semantic SLAM was able to close the loop with \gls{slc}. 
In the three-floor experiment (b), the robot took off and landed at the same position, and traveled across the entire three-story building.
Before \gls{slc}, the final pose estimates were far away from the start pose, and the chairs were reconstructed at different altitudes. The \gls{slc} was able to correct the pose estimation drift and close the loop. 
Such drastic drift correction was backward propagated in the semantic factor graph to correct robot poses and the semantic map, which is illustrated in the zoomed-in views (red boxes). 
After \gls{slc}, the robot poses and chairs were at the same altitude with the ground plane as expected.

Next, we quantitatively compared our system against the commercial VIO solution on position and yaw estimation errors. As shown in Table.~\ref{tab:benchmark}, in the three measurements from our experiments, the position estimates from the VIO system produced errors up to 1.4\% of total trajectory length. With the \gls{slc} happening at the end of each experiment, the position errors were reduced by 83.78-92.68\% and the yaw errors were reduced by 39.79-61.82\%. These measurements demonstrate the performance of our \gls{slc} algorithms.
Such drastic drift reduction is critical for the robot to construct high-fidelity maps as well as navigate safely and accurately.
We refer the reader to our demo video for more animation on \gls{slc}.

\subsection{Autonomous Exploration and Metric-Semantic Mapping}
\label{subsubsec: Exploration and Metric-Semantic Mapping}

To evaluate the effectiveness and robustness of our proposed system, the robot performed autonomous exploration missions in the multi-floor indoor environment. \cref{fig:title-figure-platform} (d-e) shows the final metric-semantic map constructed from these experiments. Our system is able to explore the environment autonomously and generate 3D maps that contain not only geometric but also semantic information about the environment. Although we are only concerned about one specific class of semantic objects (in this case, chairs) in these experiments, our algorithm can directly work with any other classes of objects that can be detected by the instance segmentation model. 

Next, we will quantitatively analyze the uncertainty reduction achieved by our proposed system. We employ the average uncertainty, in terms of the trace of covariance matrices, of robot poses and semantic maps (i.e. semantic landmarks) as evaluation metrics. The results of one autonomous exploration experiment are shown in Fig.~\ref{fig:Uncertainty without (left) and with (right) active loop closure}. As the robot explored the environment, the uncertainty of robot poses and the semantic map gradually increased. When new semantic landmarks were observed, the uncertainty surged. The uncertainty of landmarks decreased as more observations accumulated. During the exploration, the robot actively navigated to establish \gls{slc}. Upon \gls{slc}, the pose uncertainty droped sharply by 56.67\%. The subsequent observations of landmarks further reduced the uncertainty of landmarks as shown in the bottom right panel. 
We compared the results with and without the \gls{slc} module, by turning off the \gls{slc} in the latter. 
At the end of the exploration, the average robot pose uncertainty was reduced by 52.06\%, and the average landmark uncertainty was reduced by 68.53\%. 
The total trajectory length of this mission was 227.47m, where \gls{slc} happened in the middle of the mission when the robot traveled 85.6m. 
Results show that the \gls{slc} module reduced position errors by 17.07\% and yaw errors by 74.46\%. 

To demonstrate the robustness of our proposed system, the results of multiple autonomous exploration experiments are presented in Table.~\ref{tab:benchmark_reduction}. 
The second column shows the uncertainty reduction upon \gls{slc}, which is calculated based on the difference in uncertainties before and after the \gls{slc} event. 
The \gls{slc} effectively reduced the average uncertainty of robot pose by 46-57\%.
It is worth noting that, in instances where consecutive \glspl{slc} took place over a short travel distance, there was a diminishing return in terms of uncertainty reduction, as expected.
The third and fourth columns show the uncertainty reduction of poses and semantic maps, which is derived based on the difference in uncertainties with and without \gls{slc} module after the entire mission. This is achieved by simply turning on and off the \gls{slc} module.
Results demonstrate that our system achieves up to 71\% and 69\% reduction in uncertainties of robot poses and semantic maps.
The consistent reduction of errors and uncertainties in robot poses and semantic maps indicates that our system is robust and effective.

\subsection{Benchmark}
\label{subsubsec: bechmark}
\revise{Finally, we conduct benchmark experiments with state-of-the-art SLAM methods. Due to drastic viewpoint changes in our datasets, Kimera\cite{rosinol2021kimera} and ORB-SLAM3\cite{Campos_2021_orbslam} cannot detect loop closure as they require matching image features between keyframes. In addition, within the six evaluated benchmark datasets, both Kimera and ORB-SLAM3 encountered failures on certain ones. On the datasets where they both succeeded, Kimera and ORB-SLAM3 result in an odometry drift of 1.93-3.71\% and 0.45-1.51\%, respectively. However, our \gls{slc} algorithm is robust to viewpoint changes, contributing to the superior performance of our system. As a result, our system has an odometry drift that is consistently under 0.5\%.}

    \section{Conclusion}
\label{sec:conclusion}

In this paper, we developed a system for 3D exploration and metric-semantic mapping of GPS-denied indoor environments with autonomous UAVs. 
Our system features core algorithms, including metric-semantic SLAM, COP-based exploration planning, active SLC, and active uncertainty reduction planning. 
It leverages the abstractions of the environment, including exploration viewpoints extracted from the metric map, and the sparse semantic map, to significantly reduce computational load for real-time exploration and active localization.
Through extensive real-world experiments, we show the effectiveness of our proposed system in enabling the UAV to plan long-horizon paths, trading off exploration and exploitation.
Qualitative results demonstrate that our system empowered the UAV to not only explore the multi-floor environment and construct metric-semantic maps, but also intermittently establish SLC to improve the quality of the map.
The quantitative evaluation shows that our SLC module can help the robot significantly reduce position and orientation estimation errors and uncertainties. 
We envision that such a system can be deployed to solve various real-world problems. 

\bibliography{icra23ref}

\end{document}